# Miniature Fibre-Optic based Shape Sensing for Robotic Applications using Curved Reflectors

Dalia Osman (Brunel University London), Vignesh Vignesh (Brunel University London),

*Yohan Noh (Brunel University London)

*Corresponding author (yohan.noh@brunel.ac.uk)

Miniature integratable shape-sensing is crucial for precise control and measurement of curvatures made by continuum-robots, prosthetic devices and wearable body-shape sensors. A miniaturised one-degree-of-freedom joint-angle sensor is devised, using a single light emitting/receiving optical fibre with a coupler connected to a Keyence (FS-N11MN, Osaka, Japan) sensor that supplies and detects light through the optical fibres. A curvature-varying reflective surface integrated in the joint demonstrates non-contact light intensity-based sensing. Various reflector geometries and surface colours are designed to compare sensor output for achieving a large angle range and improved sensitivity for the proposed miniaturised robotic shape-sensing applications.

## 1. Introduction

The development of miniature joint angle sensors is a crucial factor for the successful utilisation of various robotic applications in the healthcare and many other industries. This includes applications such as continuum robots used in minimally invasive surgery (MIS), prosthetics, wearable flexible devices, and many more [1]. Joint angle sensing in these applications, or more broadly, shape sensing, is required to accurately actuate and measure tip position and curvatures made by these robotic devices. To do this, a number of miniaturised joint angle sensors have been developed for integration into these applications, utilising various sensor types. Some examples include inertial, stretch, and FBG-based sensors [2]. Lapusan et al [3] developed a hyper-redundant robot fused with a network of IMU sensors along the joints of the robot structure and used these to reconstruct the robot shape and position. These sensors were however relatively large for application into robotic devices that are soft or have more dexterous requirements. Alternatively, various stretch sensors have been developed for integration in soft continuum and wearable devices, such as in [4]. Here a set of carbon composite silicone elastomer films were fabricated and integrated into a robotic prosthetic hand for finger joint measurement. The bending angle is measured by the change in resistance due to stretching or change in strain. While thin and versatile, stretch sensors' materials can degrade over time, possess inhomogeneity, and experience high hysteresis in the sensor response, leading to sensing error. FBGs are highly utilised strain-based sensors, such as in [5] who developed a twisted FBG fibre for integration into soft robotic and endoscopic devices. FBG fibres are exceptionally thin and flexible, although are tricky to manufacture, leading to high costs, and exhibit increased error with higher or complex bending curvatures [6].

Alternative miniature sensor types include optical light intensity-based sensors, for example fibre optics or optoelectronic sensors. [7] presented the design of a prototype of a fibre optic based joint angle sensor for a snake-like robot. Here, five multimodal optical fibres were set on the lower link. One fibre was used to emit light, which was reflected from the flat surface of the upper link, and this reflected light was detected by the four remaining optical fibres and recorded by a set of phototransistors. These values were used estimate the joint angle, for a range of up to 35° bending angle, with a voltage range of around 4.3 V. In adding a further number of links, the number of fibre optics added could potentially increase the stiffness of the robotic structure. Alternatively, work by He et al [8] demonstrates a multisegmented structure for exoskeletal application integrated with five optoelectronic sensors. Here, the surface of each segment in the structure is utilised as a reflector for the integrated optoelectronic sensors; as the segments bend, the sensor values are used to estimate each segment bending angle, for a range of up to 80° with a voltage range of around 4.3 V, but this range is achieved using many numbers of sensors. Palli et al [9] developed a joint angle sensor for a tendon robot, where light from the LED was transmitted to the phototransistor (PT) through a channel of varying width within the joint, for an intensity modulated signal. Here, joint angle was measured within a range of up to 100°, over 2.5V. As described, due to the miniaturised size of these sensing systems, the sensing angle range and sensor sensitivity in terms of sensor voltage range is often affected.

The authors have therefore previously worked to develop a miniature joint angle sensor with improved sensitivity and sensing range [10][11]. Here, an optoelectronic sensor, comprising an LED and PT, is integrated into the joints of a robotic structure, in which a curved reflector is set around the joint. By specially altering the geometry of the reflector, allowing it to gradually change in curvature during rotation of a joint, the optoelectronic sensor was able to detect a proximity varying voltage signal. Results demonstrated better sensitivity over a larger joint angle sensing range (0-140° over a voltage range of 0-3.5 V), compared to the use of a flat joint reflector (0-40° angle sensing range over 0 – 4.3V), as was done in [12]. It was demonstrated that the sensor output could be tailored by altering the geometry parameters of the reflector curvature.

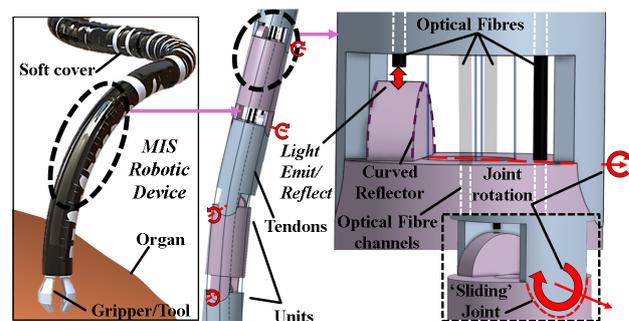

Fig. 1 Concept image for integration of fibre optic based joint angle sensor for robotic shape sensing application

To develop this idea further, this paper presents a similar premise, using a fibre, a coupler, an LED, a light detector (photodiode or PT) with the curved reflector for joint angle sensing, as presented in Figure 1. The light emitting from the LED is transmitted through the coupler, and the light reflected by the reflector is transmitted through the coupler to the light detector. Two fibres for emitting and receiving light are connected to the coupler, and they are remotely installed, so only one fibre for emitting and receiving light is installed inside the manipulator (Figure 2). This approach allows a small number of optical fibres to be embedded into the manipulator, not affecting the stiffness or size of the manipulator.

This technique combines the advantages of the improved sensor output in using a curved reflector, along with the miniscule dimensions of an optical fibre. Compared to the previously mentioned sensor types, optical fibres are low in cost and flexible. They employ non-contact light intensity-based proximity sensing seeing as they do not directly measure quantities such as strain. This means the sensors will not suffer from material inhomogeneity or load limitations, ensuring maintained robustness, and will not suffer from excessive electromagnetic interference.

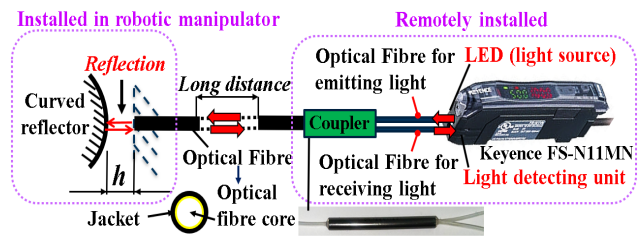

Fig.2 Schematic illustrating light intensity proximity based sensing using curved reflector and fibre optics.

## 2. Optical Fibre Sensor design

### 2.1 Sensing Principle

To develop this sensor, fibre optic technology using a coupler including optical fibres (1x2 50:50 coupler, multi-mode, 650nm, LASER 2000 Ltd., Germany) and a Keyence sensor as a light source and a light detecting unit were used. The working principle is illustrated in Figure 2. Here, two optical fibres are connected to the ports of a Keyence sensor device (FS-N11MN, Osaka, Japan); one to the LED port for emitting light, and one to the phototransistor (PT) port, for detecting reflected light. The pair of optical fibres are connected with a coupler into a single optical fibre, which is integrated into the robotic device [13]. There, light is emitted and reflected into the one fibre. The fibre optic coupler has a 50:50 coupling ratio, so half the input power is detected by the PT. The proximity ($h$) of the optical fibre to the curvature varying reflector changes during joint rotation, hence modulating the amount of light received into the optical fibre. This detected light measured by the PT can be used to estimate the joint angle. In using this sensing principle and design, the integration into robotic application becomes very simple, and easy to miniaturise, as the optical fibre is thin (900 μm), long, and flexible. The coupler and Keyence sensor can be installed remotely, away from the robotic device, further enabling reduction in the manipulator size.

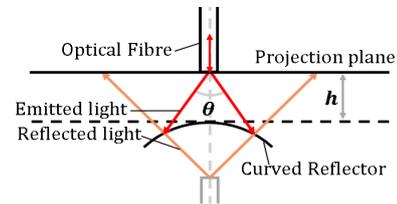

Fig. 3 Light Intensity schematic of fibre optic and curved reflector configuration

### 2.2 Light Intensity Model

This sensing principle is based on the modulation of emitted and reflected light intensity. A derived light intensity model can be used to predict the sensor behaviour under various geometrical parameters; this is given by Equation 1, based on the schematic illustrated in Figure 3 [10], [14], [15]. Here, light is transmitted from the optical fibre with angle $\theta$ onto the curved reflective surface, and reflected back onto the projection plane where the light is received by the same optical fibre. The curved reflector is set at distance $h$ from the projection plane, this varies during joint rotation. In Equation 1, the light intensity is calculated using a gaussian based model. Due to the miniscule size of the optical fibre compared to the illuminated area on the curved surface, this surface is taken to be flat. Therefore, the light intensity flux $\phi_c$ received by the fibre can be calculated as in Equation 1, where the initial intensity $I_0$ is integrated with the cross-sectional area of the optical fibre (with diameter $d$) over the reflected beam of light. The beam radius is given by $r$, and gaussian width $w$ is a function of the distance $h$ and transmitting angle $\theta$. From this, the theoretical voltage value $V_{th}$ can be estimated, by multiplying the flux by the reflection rate $R$, and conversion coefficient $k_v$ (Equation 2).

$$\phi_c = \int_{-\frac{d}{2}}^{\frac{d}{2}} I_0 \cdot e^{-2\left(\frac{r^2}{w^2}\right)} \cdot 2\pi r \, dr \qquad w = f(h,\theta) \quad (1)$$

$$V_{th} = \phi_c \cdot R \cdot k_v \qquad (2)$$

## 2.3 Curved Reflector Design

The reflector design is shown in Figure 4. As can be seen, the thickness, or curvature, of the surface gradually changes around its circumference. When mounted onto a joint, with the optical fibre opposing the surface of the reflector, the proximity between these will vary as the joint rotates. Various geometries have been designed, and these are tested using the experimental platform described in Section 3. The thickness of the curved reflectors vary from 1-2 mm, up to 1-5 mm, over angle α ranging between 120°-180°. The reflector components were 3D printed in high resolution using white resin for a smoother surface. The reflectors were tested using the bare white resin surface, and were covered in silver aluminium reflective tape, as well as white tape, to compare sensor performance.

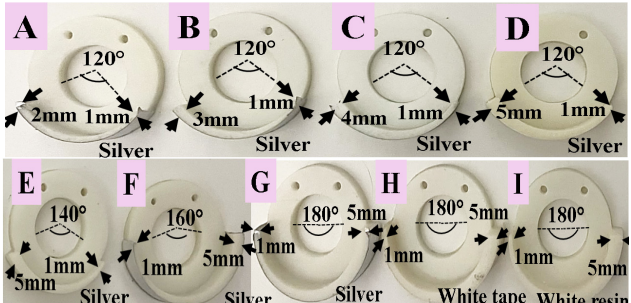

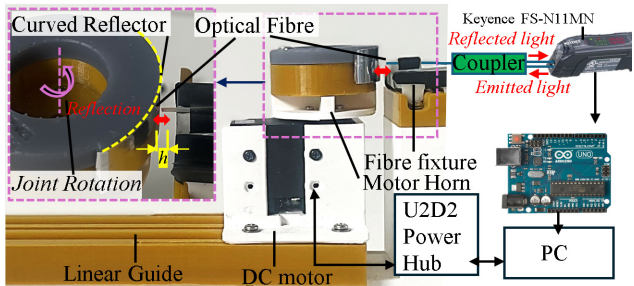

Fig.4 3D printed curved reflector prototypes for testing varying geometries and surface properties.

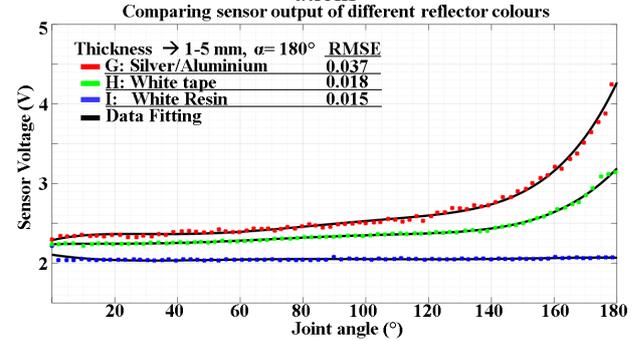

Fig.5 Experimental set up for testing sensor output with various geometrical designs of curved reflectors.

## 3. Experimental Method

The experimental platform is shown in Figure 5. Here, a one degree of freedom joint is designed. This comprises a curved reflector attached to the motor horn of a 12 V DC servo motor (Dynamixel XL430-W250T, ROBOTIS, South Korea). This motor is connected to a U2D2 power hub (ROBOTIS) to control fine joint angle motion via control software developed in Python. The motor, motor horn, and reflectors are mounted onto a linear guide platform, for fine adjustment of position relative to the optical fibre fixed on one end. The single optical fibre is connected to the coupler that joins two optical fibres, one connected to a light source (LED) and one for receiving reflected light (PT) on the Keyence sensor device ports, as shown in Figure 5. An Arduino Uno board (5V) is connected to the Keyence sensor to receive analogue sensor voltage values, as the joint angle varies. These data are collected synchronously by the PC control software. To collect all experiment data, this is repeated for each reflector shown in Figure 4. The results are shown in the following section.

## 4. Results and Analysis

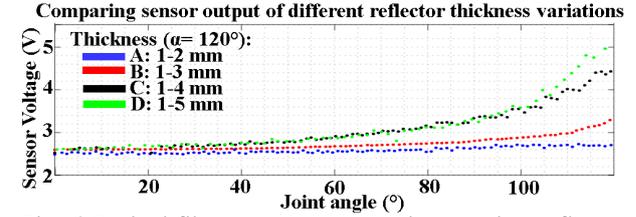

Fig. 6 Optical fibre sensor output using varying reflector thickness variations, with silver colour surface

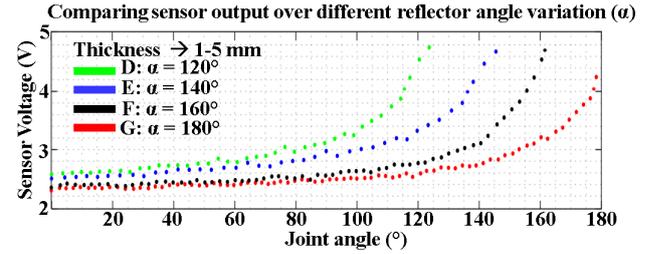

Fig.7 Sensor output compared using various angular variations

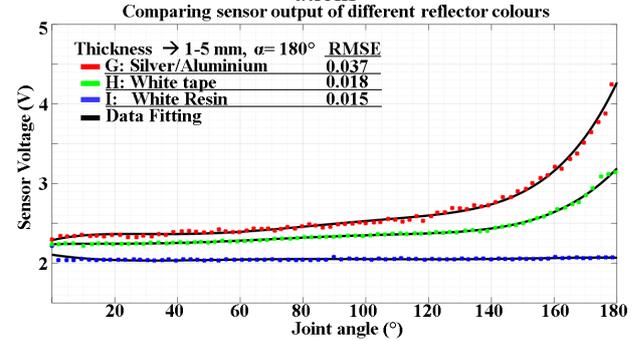

Fig. 8 Sensor output using alternate reflector colours

Figure 6 shows the optical fibre sensor output of a set of designed reflectors depicted in Figure 4 (A-D). These reflectors had varying thickness, between 1-2 mm up to 1-5mm, all varying over an angular range (α) of 120°, each covered with silver aluminium tape. It can be seen that the reflector with smaller thickness variation does not elicit a high variation voltage response, compared to one with higher thickness variation, reaching a voltage variation of around 2.5V. The sensor output displays a non-linear response, which is in line with the gaussian light intensity characteristics of the optical fibre sensor, especially with decreasing proximity to the reflector (increasing joint angle), as seen in line D in Figure 6, although has higher voltage variation. Figure 7 demonstrated the sensor response when the angle variation of the reflector (α) changes. Here, the thickness change is fixed to 1-5mm, with each reflector varying in the angular range (α) in which this thickness change occurs (D-G). Again, the surface was covered with reflective aluminium tape. Here, it can be seen how this 'α' parameter affects the sensor response. With a more gradual change in thickness from 1-5mm, over α=180° (line G, Figure 7), this gives an initial slowly rising linear response, up to around 100°, then a steeper, more nonlinear response onwards. Whereas a more drastic change in thickness from 1-5mm, over just 120° (Line D, Figure 7), elicits a more imminent, steeper response, although this reaches maximum voltage of around 4.5 V at the 120° joint angle. As such, these parameters can be altered to achieve the required sensor response for certain tasks. In looking at line F, Figure 7, for example, we can extract a usable sensor response from the joint angle of 80° onwards, up to 160°. This gives a sensing range of 80° with voltage variation of around 2V. Comparing this with some of the ranges given by joint angle sensors mentioned in the introduction, this shows potential for further development as a simple solution for joint angle sensing in a variety of soft and flexible robotic applications. Figure 8 illustrates how the reflectivity of the surface of the reflector can affect the sensor response, where identical reflectors (G-I) are tested with the optical fibre over a set of joint angles. As seen, the silver aluminium tape surface with the highest reflectivity elicits the strongest response, compared to the response given by the plain white resin surface. Additionally, a fifth order polynomial function was used to fit the data, shown in black (Figure 8), with RMSE values shown, at a maximum of 0.037° error, verifying that this can be used to estimate joint angle based on sensor voltage data with good accuracy. The signal measured through these experiments did exhibit some level of noise, which should be targeted by introducing shielding of ambient light. Another source of noise may have also been due to any irregularities in the reflector surface texture during the application of the tape. Therefore, future prototypes should be made from metallic or materials of the intended colour with precise manufacturing techniques at a high resolution. Furthermore, the voltage variation achieved during these tests were between 2.5V – 5V, due to the use of the Keyence Sensor; although through the use of an LED and Photodiode, along with an amplifier and lowpass filter, the resolution can be improved.

## 5. Conclusion

In this paper we have demonstrated the development of a simple low-cost miniature joint angle sensor utilising a single optical fibre sensor, with geometrically curved reflectors for targeting improved sensor response. The sensor allows measurement of a large range of joint angles without the need of multiple sensors, or optical fibres that affect flexibility of the device. The light-intensity based sensor is versatile for many applications, as it does not suffer from electromagnetic interference or material inhomogeneities. Future work will target more developed prototypes to demonstrate full shape sensing capabilities.

## ACKNOWLEDGEMENT

This research has received funding from MSc project 2022-2023, Department of Mechanical and Aerospace Engineering, Brunel University London. The Great Britain Sasakawa Foundation has supported attendance to the 42nd annual conference of the robotics society of Japan.